\pdfoutput=1

\documentclass[11pt]{article}

\usepackage{emnlp2022}

\usepackage{times}
\usepackage{latexsym}
\usepackage{booktabs}
\usepackage{graphicx}
\usepackage{tabularx}
\usepackage{svg}
\usepackage{relsize}
\usepackage{amsmath}
\usepackage{multirow}
\usepackage{tikz}
\usepackage{adjustbox}
\usepackage{colortbl}
\usepackage{enumitem}
\usepackage{todonotes}
\usepackage{amssymb}
\usepackage{pifont}

\usepackage[T1]{fontenc}

\usepackage[utf8]{inputenc}

\usepackage{microtype}

\usepackage{inconsolata}
\usepackage{textcomp}

\newcommand{\nmtscore}{\mbox{\textsc{NMTScore}}}
\newcommand{\nmtsc}{\mbox{\textsc{NMTSc.}}}
\newcommand{\nmtscoredirect}{\nmtscore{}$\mkern1mu\text{-}\mkern1mu\text{direct}$}
\newcommand{\nmtscorepivot}{\nmtscore{}$\mkern1mu\text{-}\mkern1mu\text{pivot}$}
\newcommand{\nmtscorecross}{\nmtscore{}$\mkern1mu\text{-}\mkern1mu\text{cross}$}

\newcommand{\nmtscdir}{\nmtsc{}$\text{-}\text{dir.}$}
\newcommand{\nmtscpivot}{\nmtsc{}$\text{-}\text{pivot}$}

\newcommand{\hypref}{$\textrm{hyp}\mkern1mu|\mkern1mu\textrm{ref}$}
\newcommand{\refhyp}{$\textrm{ref}\mkern1mu|\mkern1mu\textrm{hyp}$}

\newcommand{\chrf}{\mbox{\textsc{chrF}}}
\newcommand{\bertscore}{\mbox{\textsc{BERTScore}}}
\newcommand{\sentbleu}{\mbox{\textsc{sentBLEU}}}
\newcommand{\sentencebert}{\mbox{Sentence-BERT}}
\newcommand{\xlmroberta}{\mbox{XLM-RoBERTa}}
\newcommand{\roberta}{\mbox{RoBERTa}}
\newcommand{\pawsx}{\mbox{PAWS-X}}
\newcommand{\prism}{\textit{Prism}}

\newcommand{\ExternalLink}{%
    \tikz[x=1.2ex, y=1.2ex, baseline=-0.05ex]{%
        \begin{scope}[x=1ex, y=1ex]
            \clip (-0.1,-0.1)
                --++ (-0, 1.2)
                --++ (0.6, 0)
                --++ (0, -0.6)
                --++ (0.6, 0)
                --++ (0, -1);
            \path[draw,
                line width = 0.5,
                rounded corners=0.5]
                (0,0) rectangle (1,1);
        \end{scope}
        \path[draw, line width = 0.5] (0.5, 0.5)
            -- (1, 1);
        \path[draw, line width = 0.5] (0.6, 1)
            -- (1, 1) -- (1, 0.6);
        }
    }

\title{\nmtscore{}: A Multilingual Analysis of \\ Translation-based Text Similarity Measures}

\author{Jannis Vamvas$^1$ \and Rico Sennrich$^{1,2}$\\
  $^1$Department of Computational Linguistics, University of Zurich\\
  $^2$School of Informatics, University of Edinburgh \\ \medskip
  \texttt{\{vamvas,sennrich\}@cl.uzh.ch}}

\begin{document}
\maketitle
\begin{abstract}
Being able to rank the similarity of short text segments is an interesting bonus feature of neural machine translation.
Translation-based similarity measures include direct and pivot translation probability, as well as \textit{translation cross-likelihood}, which has not been studied so far.
We analyze these measures in the common framework of multilingual NMT, releasing the \nmtscore{} library.
Compared to baselines such as sentence embeddings, translation-based measures prove competitive in paraphrase identification and are more robust against adversarial or multilingual input, especially if proper normalization is applied.
When used for reference-based evaluation of data-to-text generation in 2~tasks and 17~languages, translation-based measures show a relatively high correlation to human judgments.
\end{abstract}

\section{Introduction}\label{sec:introduction}

Measures of paraphrastic similarity aim to quantify the degree to which text segments mean the same thing.
Such measures can be used to identify paraphrases, and also to automatically evaluate text generation by estimating the similarity of model outputs to human-written references.

Neural machine translation~(NMT) enables several similarity measures as a by-product of learning to estimate the probability of translations~\cite{mallinson-etal-2017-paraphrasing,junczys-dowmunt-2018-dual,thompson-post-2020-automatic}.
These measures are promising given that they naturally leverage parallel corpora and might pay more attention to details such as word order or named entities than sentence embeddings do.
For example, \citet{zhang-etal-2019-paws} have demonstrated that it is difficult to spot the mismatch between \textit{``Flights from New York to Florida''} and \textit{``Flights from Florida to New York''} purely based on pooled representations, and they have released a challenge set of such paraphrase adversaries called PAWS.

\begin{figure}[t]
  \centering
  \includegraphics[width=\linewidth]{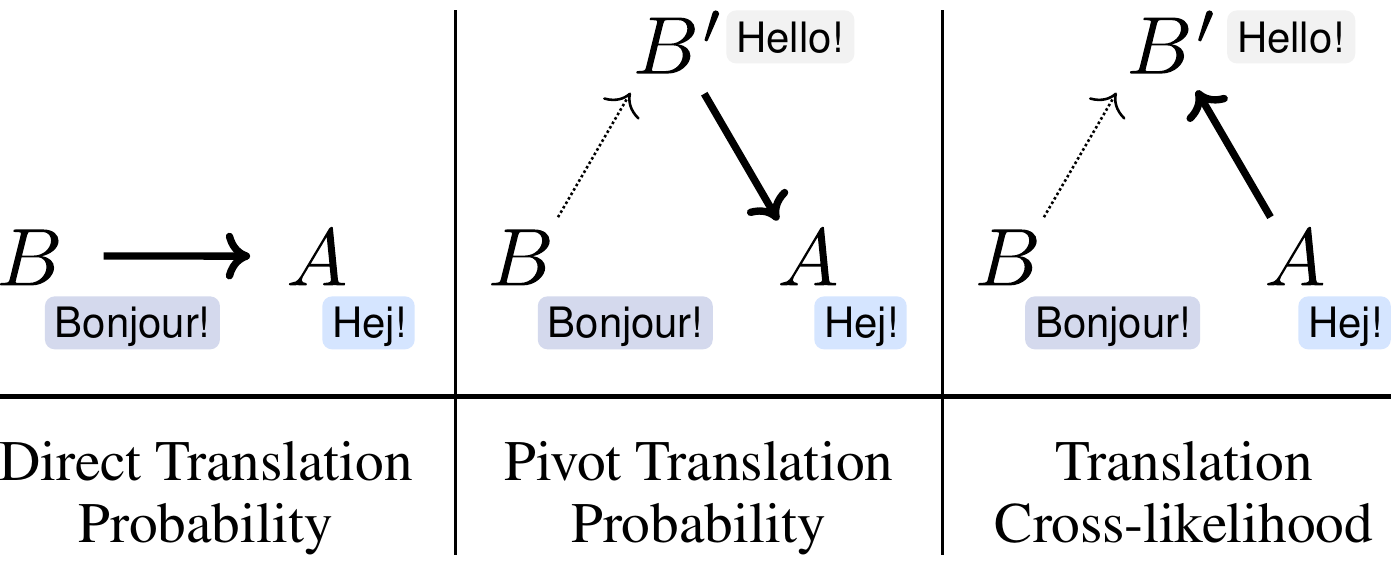}
  \caption{
  \label{fig:figure-1}
  Three text similarity measures are analyzed in this paper.
  Each measure uses a specific translation probability (thick arrow) to estimate the similarity of sentence~$A$ to~$B$.
  Some measures require a translation~$B'$ into an intermediate language (thin arrow).
  }
\end{figure}

In this paper, we consider three distinct subtypes of translation-based similarity, as visualized in Figure~\ref{fig:figure-1}.
First of all, a straightforward approach is to estimate the \textit{direct translation probability} of sentence~$A$ when translating from sentence~$B$~\cite{junczys-dowmunt-2018-dual, thompson-post-2020-automatic}.
Secondly, \textit{pivot translation probability} is an estimate of how probable it is to arrive at sentence~$A$ when pivoting through an intermediary language~\cite{mallinson-etal-2017-paraphrasing}.
Finally, we propose to estimate \textit{translation cross-likelihood}: the likelihood that a translation of~$B$ into some language could also be a translation of~$A$.

We release the \nmtscore{} library, using many-to-many multilingual NMT to implement these measures.\footnote{\url{https://github.com/ZurichNLP/nmtscore}}
All three measures are competitive in paraphrase identification across 9~languages, compared to other general-purpose similarity measures, and especially on adversarial examples.
We find, however, that normalizing the measures with reconstruction probability is important.
They also work well cross-lingually, where pivot translation probability performs best while cross-likelihood has the advantage of not requiring an explicit specification of the input's languages.

With respect to reference-based evaluation, we show that \nmtscore{} is a competitive evaluation metric for data-to-text, according to human judgments from the WebNLG Challenge~\cite{ferreira2020webnlg} and from a multilingual AMR-to-text evaluation~\cite{fan-gardent-2020-multilingual}.
Taken together, multilingual NMT offers a relatively precise and often complementary perspective on paraphrastic similarity, with little correlation to other metrics.

In summary, we make the following main contributions:
\begin{itemize}[itemsep=1pt]
    \item We redefine translation-based similarity measures in a common multilingual framework, proposing \textit{reconstruction normalization} and a novel \textit{translation cross-likelihood} measure.
    \item We compile a multilingual paraphrase identification benchmark, showing that \nmtscore{} outperforms other general-purpose measures.
    \item We demonstrate that \nmtscore{} provides effective metrics for reference-based evaluation of data-to-text generation.
\end{itemize}

\section{Translation-based Similarity Measures}\label{sec:translation-based-similarity-measures}

The similarity of two sentences~$A$ and~$B$ can be measured in several ways by using a multilingual translation model~$\theta$.
Such a model accepts multiple source languages and can translate into multiple target languages.
Usually only the target language needs to be specified, e.g.\ with a target language token, and thus we use $\theta_{\ell}$ to denote a model that is conditioned on a target language $\ell$.

\subsection{Direct Translation Probability}\label{subsec:direct-translation-probability}

\paragraph{Cross-lingual}
If $A$ and $B$ are in two different languages $a$ and $b$, the model can directly estimate the translation probability of $A$ given $B$:
\[\mathrm{P}_{\mathrm{direct}}(A|B)=p_{\theta_{a}}(A|B)\]
This probability is sometimes called the \textit{translational equivalence} of~$A$ and~$B$.
In practice, there are different ways how such a probability can be calculated from the token-level probabilities predicted by the model.
For this and the following measures, we follow previous work~\cite{junczys-dowmunt-2018-dual,thompson-post-2020-automatic} and normalize by sequence length:
\[p_{\theta_{a}}(A|B) := \Big[ \prod_{i=0}^{|A|}p_{\theta_{a}}(A^{i}|B,A^{<i})\Big]^\frac{1}{|A|}\]

\paragraph{Monolingual}
Since~$\theta$ is a multilingual model,~$A$ and~$B$ may also be in the same language~\cite{thompson-post-2020-automatic}.
This is because multilingual NMT enables zero-shot translation~\cite{johnson-2017-google}, which includes any monolingual direction~$\ell\rightarrow{}\ell$.
\citet{thompson-post-2020-automatic} argue that monolingual translation can be seen as (non-diverse) paraphrasing, and they have demonstrated that paraphrasing probability is a useful metric for reference-based MT evaluation (called \prism{}).

\subsection{Pivot Translation Probability}\label{subsec:pivot-translation-probability}
Paraphrastic similarity can also be estimated via translation to a pivot language~\cite{bannard-callison-burch-2005-paraphrasing, mallinson-etal-2017-paraphrasing}.
Pivot translation requires two translation directions, $\theta_{pivot}$ and~$\theta_{a}$.
First, a translation $B' \sim p_{\theta_{pivot}}(\cdot|B)$ is generated and then used to calculate the probability of translating $B'$ into $A$:
\[\mathrm{P}_{\mathrm{pivot}}(A|B)=p_{\theta_{a}}(A|B')\]
Such an approach is typically used for monolingual sentences (\textit{round-trip translation}), but we argue that~$A$ and~$B$ can also be in two different languages.
Furthermore, the pivot language may be identical to the language of~$A$ or~$B$, considering the zero-shot paraphrasing capability of multilingual NMT.

\subsection{Translation Cross-likelihood}\label{subsec:similarity-of-conditionals}
As an alternative measure we propose \textit{translation cross-likelihood}, which requires only one translation direction~$\theta_{tgt}$, where~$tgt$ is any target language supported by the NMT model.
We generate a translation $B' \sim p_{\theta_{tgt}}(\cdot|B)$ and then estimate the likelihood that~$B'$ could have been generated from~$A$:
\[\text{Cross-likelihood}(A|B)=p_{\theta_{tgt}}(B'|A)\]
In other words, this similarity reflects the surprisal of a translation model that is conditioned on sentence~$A$ but exposed to a translation of sentence~$B$.

Like with pivot translation,~$A$ and~$B$ may be a monolingual or a cross-lingual pair, and the target language may or may not be identical to the language of~$A$ and~$B$.

\subsection{From Probability to Similarity Measure}\label{subsec:from-probability-to-similarity-score}

\paragraph{Normalization}
Similarity measures typically assign maximum similarity to indiscernible inputs.
We propose to ensure this by applying the following normalizations to translation-based measures:
\begin{alignat*}{2}
&\nmtscore{}\mkern1mu\text{-}\mkern1mu\text{direct}(A|B) &&= \frac{p_{\theta_{a}}(A|B)}{p_{\theta_{a}}(A|A)} \\[2ex]
&\nmtscore{}\mkern1mu\text{-}\mkern1mu\text{pivot}(A|B) &&= \frac{p_{\theta_{a}}(A|B')}{p_{\theta_{a}}(A|A')} \\[2ex]
&\nmtscore{}\mkern1mu\text{-}\mkern1mu\text{cross}(A|B) &&= \frac{p_{\theta_{tgt}}(B'|A)}{p_{\theta_{tgt}}(B'|B)}
\end{alignat*}
The two formulas for \nmtscoredirect{} and \nmtscorepivot{} can be seen as a form of \textit{reconstruction normalization}, given that
 $p_{\theta_{a}}(A|A)$ is the probability that the sentence remains identical when zero-shot paraphrasing is performed.
Likewise, $p_{\theta_{a}}(A|A')$ is the pivot reconstruction probability of~$A$ given its pivot translation $A' \sim p_{\theta_{pivot}}(\cdot|A)$.
The latter measure has also been used by~\citet{mallinson-etal-2017-paraphrasing} for normalization.

\paragraph{Symmetrization}
Translation probabilities are directed measures, however the order of $A$ and $B$ is often arbitrary, such as in paraphrase identification.
We thus follow previous work~\cite{junczys-dowmunt-2018-dual,thompson-post-2020-automatic} and average the directed measures for both directions:
\[sim(A, B) = \frac{1}{2}\,sim(A | B) + \frac{1}{2}\,sim(B | A)\]

\section{Baseline Measures}\label{sec:baseline-measures}

\subsection{Surface Similarity Measures}\label{subsec:surface-similarity-measures}

Some well-known text similarity measures, especially for reference-based evaluation, rely on surface similarity.
We choose \chrf{}~\cite{popovic-2015-chrf} and sentence-level BLEU~\cite{papineni-etal-2002-bleu} as surface-similarity baselines.
\chrf{} is a character-based metric that calculates precision and recall of character \textit{n}-grams.
BLEU calculates the precision of word \textit{n}-grams with a brevity penalty.

\subsection{Embedding-based Similarity Measures}\label{subsec:embedding-based-similarity-measures}
Another family of similarity measures uses the cosine similarity of text embeddings.
Such embeddings are typically learned on the token level, and thus need to be aggregated in some way.
In this paper, we consider two embedding baselines:

\paragraph{Similarity of aggregate token embeddings}
A typical approach is to average the token embeddings before calculating cosine similarity.
However, hidden states of self-supervised Transformer language models may not be directly useful when averaged;
\citet{reimers-gurevych-2019-sentence} fine-tune them using a sentence pair classification objective, calling their approach \sentencebert{}.

\paragraph{Aggregation of token similarities}
An alternative is to aggregate the similarities between the individual tokens of the two sentences~\cite{mihalcea2006corpus}.
Typically, a \textit{precision} is calculated as the average maximum cosine similarity of all tokens in $A$ to all tokens in $B$, and a \textit{recall} is calculated with $A$ and $B$ switched.
It has been shown that when aggregated in this way, hidden states of self-supervised language models are useful for text similarity even without any fine-tuning~\cite{mathur-etal-2019-putting,zhang2019bertscore}.
This measure is called \bertscore{} by \citet{zhang2019bertscore}.

\section{Paraphrase Identification}\label{sec:paraphrase-identification}

In this section we compare the similarity measures using paraphrase identification test sets in multiple languages.
The test sets contain pairs of sentences that have been annotated with whether the sentences are paraphrases or not, yielding a binary classification problem.
For datasets with a validation split, we determine thresholds that optimally separate the validation set for each measure; we then apply the thresholds to the test set to compute the accuracy of the measures.
If there is no validation set, we report the Area Under the Curve~(AUC) on the test set.

\begin{table*}[]
\begin{tabularx}{\textwidth}{@{}Xllll@{\hskip 20pt}llllll@{\hskip 20pt}r@{}}
\toprule
 &
  \multicolumn{4}{l}{\mbox{\textbf{Individual datasets}}} &
  \multicolumn{5}{@{}l}{\textbf{\pawsx{} dataset}} &
  \multicolumn{2}{r@{}}{\mbox{\textbf{Macro-}}} \\
  Language &
  \textsc{en} &
  \textsc{ru} &
  \textsc{fi} &
  \textsc{sv} &
  \textsc{de} &
  \textsc{es} &
  \textsc{fr} &
  \textsc{ja} &
  \textsc{zh} &
  \multicolumn{2}{r@{}}{\mbox{\textbf{average}}} \\
  Metric &
  Acc. &
  \textsc{auc} &
  \textsc{auc} &
  \textsc{auc} &
  Acc. &
  Acc. &
  Acc. &
  Acc. &
  Acc. &
  Avg. &
   \\ \midrule
\multicolumn{12}{@{}l}{\textit{Surface similarity baselines}} \\ \addlinespace[3pt]
\chrf &
  69.9 &           
  78.4 &           
  58.9 &           
  66.8 &           
  58.3 &           
  57.2 &           
  58.5 &           
  56.2 &           
  59.3 &           
  57.9 &           
  66.4 \\          
\sentbleu &
  64.2 &           
  70.1 &           
  63.6 &           
  62.2 &           
  61.3 &           
  59.9 &           
  61.2 &           
  55.4 &           
  59.0 &           
  59.4 &           
  63.9 \\ \midrule 
\multicolumn{12}{@{}l}{\textit{Embedding baselines}} \\ \addlinespace[3pt]
\mbox{\sentencebert} &
  \textbf{71.2} &           
  83.5 &           
  68.7 &           
  \textbf{73.0} &           
  58.2 &           
  58.2 &           
  59.5 &           
  55.5 &           
  58.1 &           
  57.9 &           
  70.9 \\          
\bertscore-F1 &
  \textbf{72.6} &           
  79.7 &           
  66.6 &           
  65.9 &           
  59.8 &           
  60.2 &           
  60.5 &           
  56.7 &           
  59.5 &           
  59.4 &           
  68.8 \\ \midrule 
\multicolumn{12}{@{}l}{\textit{Translation-based measures}} \\ \addlinespace[3pt]
\nmtscoredirect &
  \textbf{72.6} &           
  84.1 &           
  \textbf{72.4} &           
  70.6 &           
  73.9 &           
  73.5 &           
  75.7 &           
  66.4 &           
  68.9 &           
  71.7 &           
  74.3 \\          
\nmtscorepivot &
  \textbf{72.1} &           
  84.9 &           
  70.3 &           
  \textbf{70.9} &           
  \textbf{77.4} &           
  \textbf{76.2} &           
  \textbf{76.9} &           
  \textbf{68.4} &           
  \textbf{70.8} &           
  \textbf{74.0} &           
  74.4 \\          
\nmtscorecross &
  \textbf{71.7} &           
  \textbf{86.6} &           
  71.2 &           
  \textbf{72.4} &           
  \textbf{76.6} &           
  \textbf{75.1} &           
  75.6 &           
  65.8 &           
  \textbf{70.5} &           
  72.7 &           
  \textbf{74.9} \\ \bottomrule 
\end{tabularx}
\caption{
Comparison of general-purpose text similarity measures on paraphrase identification in 9~languages.
While the embedding baselines build on \xlmroberta{}~\cite{conneau-etal-2020-unsupervised}, the translation-based measures use probability estimates of a multilingual NMT system~\citep[\textit{Prism,}][]{thompson-post-2020-automatic}.
We report \textsc{auc} if there is no validation set for a language.
Results within the top significance cluster are printed in bold.
}
\label{tab:paraphrase-results}
\end{table*}

\subsection{Experimental Setup}\label{subsec:experimental-setup}

\paragraph{Translation model}

We use the same multilingual NMT model for all three translation-based measures.
Specifically, we use a 745M-parameter Transformer model (\prism{}) that was trained by~\citet{thompson-post-2020-automatic} using Fairseq~\cite{ott-etal-2019-fairseq}.\footnote{\url{https://github.com/thompsonb/prism}}
The model supports 39 languages and is not English-centric.
We found no indication that its training data overlap with the datasets used in the experiments.

\paragraph{Intermediary language}
We use English as the pivot language for pivot translation and as the target language for estimating cross-likelihood.
English has the largest share of training data in the models we use.

\paragraph{Surface Similarity baselines}
We compute sentence-level \chrf{} and sentence-level BLEU using the SacreBLEU library~\cite{post-2018-call}.
We use the recommended tokenization for BLEU, tokenizing Japanese text using MeCab and splitting Chinese characters individually.
When applying \chrf{} and BLEU to paraphrase identification, we calculate the similarity in both directions and take the average.

\paragraph{Embedding baselines}
We use pre-trained embeddings from \xlmroberta{}, which is a multilingual masked language model pre-trained on CommonCrawl~\cite{conneau-etal-2020-unsupervised}.
We use the large version (550M~parameters) to compute \bertscore{}, specifically the 17th layer as recommended by the \bertscore{} reference implementation\footnote{\url{https://github.com/Tiiiger/bert_score}}.
For \sentencebert{}, we use a version of size `base' (270M~parameters) that~\citet{reimers-gurevych-2020-making} have finetuned by distilling the sentence embeddings of an English \roberta{} model.
The latter has in turn been fine-tuned on 50M English paraphrase pairs, which do not overlap with our test sets.
The distillation was performed using parallel sentences for 50~languages.

\subsection{Datasets}\label{subsec:datasets}
We use the following datasets for our experiments (statistics are reported in Appendix~\ref{sec:dataset-statistics}):

\paragraph{English} MRPC~\cite{dolan-brockett-2005-automatically}, a corpus of sentence pairs automatically extracted from news, and annotated with binary labels.
We exclude samples where a re-annotation effort by~\citet{kovatchev-etal-2018-etpc} has found inconsistent labeling.

\paragraph{Russian} ParaPhraser~\cite{pivovarova2017paraphraser}, a corpus of news headlines annotated on a three-class ordinal scale.
We follow the original setup and create binary labels by merging \textit{precise} and \textit{near} paraphrases into a single class.

\paragraph{Finnish and Swedish} The Finnish Paraphrase Corpus~\cite{kanerva-etal-2021-finnish}, a dataset of manually selected subtitle lines and news headlines, annotated on a four-class ordinal scale.
A small test set in Swedish is likewise available.
We create binary labels by categorizing all pairs with a label of~4 as positive, and all below as negative.

\begin{figure*}[t]
  \centering
  \includegraphics[width=\textwidth]{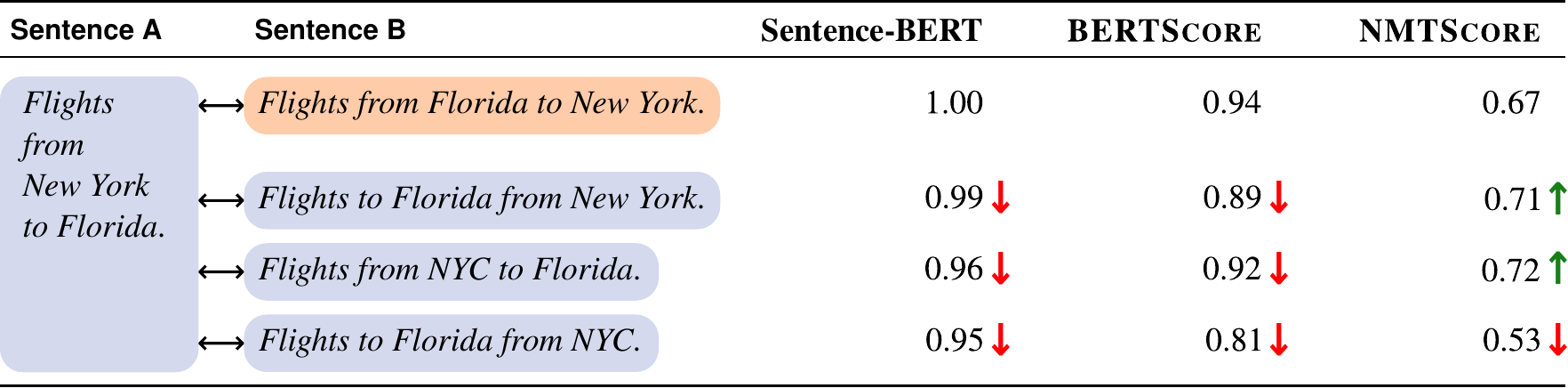}
  \caption{
  \label{fig:qualitative}
An adversarial example originally used by~\citet{zhang-etal-2019-paws}, judged by three similarity metrics: \sentencebert{} cosine similarity, rescaled \bertscore{}-F1 and \nmtscorecross{}.
The arrows indicate whether a paraphrase pair is assigned a higher or lower similarity than the pair in the first row, which is a non-paraphrase.
  }
\end{figure*}

\paragraph{\pawsx{}}
We also include \pawsx{}~\cite{yang-etal-2019-pawsx}, a challenge set of sentence pairs with high word overlap in a total of 7~languages.
The dataset is based on English sentences extracted from Wikipedia that have been paired with automatically created derivative sentences, and annotated with binary labels~\cite{zhang-etal-2019-paws}.
Test sets in other languages have been created by manually translating the English sentences.
We report results for German, Spanish, French, Japanese, Chinese and an average over these languages.
We do not report results for English, since a part of the positive examples in the original English dataset have been created with automatic round-trip translation.
The other languages are better suited for our analysis, because the manual translation process has broken the direct link between the original sentence and its round-trip translation.
We also do not report results for Korean, since the \prism{} translation model we use for our main experiments has not been trained in that language.

\subsection{Statistical Analysis}\label{subsec:statistical-analysis}

For all comparisons between the translation-based and baseline measures we broadly follow the methodology of the WMT Metrics Task~\cite{freitag-etal-2021-results}.
We perform paired bootstrap resampling~\cite{koehn-2004-statistical} with 1000 repetitions to assess the statistical significance of a difference between two measures at $\alpha=0.05$.
For each dataset, we determine the \textit{top significance cluster}, which is printed in bold.
This cluster contains the top measures that are not significantly outperformed by any other measure.
We also perform significance tests for average metrics over multiple datasets by combining the \textit{i}th bootstrap sample of every dataset into the \textit{i}th bootstrap sample of the full benchmark.
Our analysis was implemented using the SacreROUGE framework~\cite{deutsch-roth-2020-sacrerouge}.

\subsection{Monolingual Results}

The results of monolingual paraphrase identification are shown in Table~\ref{tab:paraphrase-results}.
In the final column we report the macro-average over all datasets, i.e., the average of \textsc{en} accuracy, \textsc{ru} \textsc{auc}, \textsc{fi} \textsc{auc}, \textsc{sv} \textsc{auc} and average \pawsx{} accuracy.

Overall, the translation-based measures perform better than the baselines that use surface similarity or embeddings.
They excel on the adversarial \pawsx{} dataset, with an improvement of 10--15 points over the baselines.
On the other datasets the accuracy of the translation-based measures is comparable to the embedding baselines.

It should be noted that the embedding baselines and the translation-based measures are not perfectly comparable, since multilingual models with different hyperparameters and pretraining languages are used.
The three translation-based measures are perfectly comparable, since they rely on the same NMT model.
Translation cross-likelihood has a slightly higher accuracy than direct or pivot translation probability in monolingual paraphrase detection.
Appendix~\ref{tab:m2m100-results} shows that this finding can be reproduced with an alternative multilingual NMT system in two different sizes~(\textit{M2M-100};~\citealp{fan2021beyond}).

\subsection{Qualitative Analysis}\label{subsec:qualitative-analysis}
A qualitative comparison suggests that translation-based measures are superior in distinguishing numbers, named entities and enumerations, but that embeddings can better capture the similarity of sentences with similar meaning but very different phrasing~(Appendices~\ref{sec:examples} and~\ref{sec:cross-lingual-pawsx-examples}).

 Figure~\ref{fig:qualitative} illustrates this observation on the example of the adversarial pair mentioned in the introduction.
Both \sentencebert{} and \bertscore{} assign a higher score to the non-paraphrase with a high word overlap, and a lower score to paraphrases with a difference in word order or entity naming.
\nmtscore{} accurately outputs a higher score for the two paraphrases than for the non-paraphrase but still fails on a fourth sentence pair that combines both phenomena.

\begin{table*}[]
\setlength{\tabcolsep}{3.5pt}
\begin{adjustbox}{width=\textwidth,center}
\begin{tabularx}{1.135\textwidth}{@{}Xcccccccccccccccr@{}}
\toprule
 &
  \textsc{en} &
  \textsc{en} &
  \textsc{en} &
  \textsc{en} &
  \textsc{en} &
  \textsc{de} &
  \textsc{de} &
  \textsc{de} &
  \textsc{de} &
  \textsc{es} &
  \textsc{es} &
  \textsc{es} &
  \textsc{fr} &
  \textsc{fr} &
  \textsc{ja} &
   \\ \addlinespace[-6pt]
 &
  + &
  + &
  + &
  + &
  + &
  + &
  + &
  + &
  + &
  + &
  + &
  + &
  + &
  + &
  + &
  Avg.\\ \addlinespace[-5pt]
 &
  \textsc{de} &
  \textsc{es} &
  \textsc{fr} &
  \textsc{ja} &
  \textsc{zh} &
  \textsc{es} &
  \textsc{fr} &
  \textsc{ja} &
  \textsc{zh} &
  \textsc{fr} &
  \textsc{ja} &
  \textsc{zh} &
  \textsc{ja} &
  \textsc{zh} &
  \textsc{zh} &
   \\ \midrule
\multicolumn{12}{@{}l}{\textit{Surface similarity baselines}} \\ \addlinespace[3pt]
\chrf &
  54.9 &
  54.9 &
  54.3 &
  54.6 &
  54.8 &
  54.9 &
  54.5 &
  54.5 &
  54.6 &
  54.7 &
  54.6 &
  54.8 &
  54.5 &
  54.7 &
  54.7 &
  54.7 \\
\sentbleu{} &
  56.4 &
  56.2 &
  56.3 &
  54.6 &
  54.7 &
  56.0 &
  56.2 &
  54.5 &
  54.5 &
  56.1 &
  54.6 &
  54.7 &
  54.5 &
  54.6 &
  54.7 &
  55.2 \\ \midrule
\multicolumn{12}{@{}l}{\textit{Embedding baselines}} \\ \addlinespace[3pt]
\sentencebert &
  53.9 &
  54.7 &
  54.8 &
  54.6 &
  54.7 &
  55.5 &
  56.0 &
  54.5 &
  54.5 &
  56.2 &
  54.6 &
  54.6 &
  54.5 &
  54.6 &
  54.7 &
  54.8 \\
\bertscore-F1 &
  57.4 &
  57.5 &
  57.0 &
  55.1 &
  54.8 &
  57.2 &
  57.3 &
  54.6 &
  54.5 &
  57.6 &
  54.7 &
  54.9 &
  54.6 &
  54.6 &
  55.0 &
  55.8 \\ \midrule
\multicolumn{12}{@{}l}{\textit{Translation-based measures}} \\ \addlinespace[3pt]
\nmtscoredirect{} &
  76.4 &
  \textbf{76.4} &
  76.1 &
  \textbf{68.6} &
  68.8 &
  73.3 &
  74.5 &
  66.0 &
  66.9 &
  74.3 &
  \textbf{66.7} &
  66.8 &
  \textbf{66.8} &
  67.4 &
  64.4 &
  70.2 \\
\nmtscorepivot{} &
  \textbf{77.4} &
  \textbf{76.9} &
  \textbf{77.3} &
  \textbf{68.9} &
  \textbf{70.7} &
  \textbf{75.0} &
  \textbf{76.0} &
  \textbf{67.0} &
  \textbf{69.5} &
  \textbf{75.5} &
  \textbf{67.6} &
  \textbf{69.5} &
  \textbf{67.5} &
  \textbf{69.9} &
  \textbf{66.5} &
  \textbf{71.7} \\
\nmtscorecross &
  76.0 &
  75.9 &
  75.9 &
  65.2 &
  66.0 &
  \textbf{74.5} &
  \textbf{75.2} &
  64.8 &
  65.8 &
  74.2 &
  64.6 &
  66.2 &
  64.4 &
  65.7 &
  65.3 &
  69.3 \\ \bottomrule
\end{tabularx}
\end{adjustbox}
\caption{
Comparison of text similarity measures on cross-lingual paraphrase identification using the \pawsx{} dataset.
Results within the top significance cluster are printed in bold.
}
\label{tab:cross-lingual-pawsx-results}
\end{table*}

\begin{table*}[]
\setlength{\tabcolsep}{3.5pt}
\begin{adjustbox}{width=\textwidth,center}
\begin{tabularx}{1.135\textwidth}{@{}Xp{0.135\textwidth}p{0.135\textwidth}p{0.135\textwidth}p{0.135\textwidth}p{0.135\textwidth}p{0.145\textwidth}}
\toprule
 &
  \chrf{} &
  \sentbleu{} &
  \textsc{SBERT} &
  \textsc{BERTSc.} &
  \nmtscdir &
  \nmtscpivot \\ \midrule
\sentbleu{} &
  \cellcolor[HTML]{C9DBF8}0.56$\pm$.01
  &
  &
  &
  & \\
\sentencebert &
  \cellcolor[HTML]{E6EFFC}0.42$\pm$.01 &
  \cellcolor[HTML]{F6F9FE}0.35$\pm$.01
  &
  &
  & \\
\bertscore &
  \cellcolor[HTML]{D1E1F9}0.52$\pm$.01 &
  \cellcolor[HTML]{E1EBFB}0.45$\pm$.01 &
  \cellcolor[HTML]{DFEAFB}0.46$\pm$.01
  &
  & \\
\nmtscoredirect{} &
  \cellcolor[HTML]{D8E5FA}0.49$\pm$.01 &
  \cellcolor[HTML]{E1EBFB}0.45$\pm$.01 &
  \cellcolor[HTML]{D1E0F9}0.52$\pm$.01 &
  \cellcolor[HTML]{CEDEF9}0.54$\pm$.01
  & \\
\nmtscorepivot{} &
  \cellcolor[HTML]{E2ECFB}0.44$\pm$.01 &
  \cellcolor[HTML]{EAF2FD}0.40$\pm$.01 &
  \cellcolor[HTML]{D6E4FA}0.50$\pm$.01 &
  \cellcolor[HTML]{DCE8FB}0.47$\pm$.01 &
  \cellcolor[HTML]{9EBEF2}0.77$\pm$.01
  & \\
\nmtscorecross{} &
  \cellcolor[HTML]{DCE8FB}0.47$\pm$.01 &
  \cellcolor[HTML]{E8F0FC}0.41$\pm$.01 &
  \cellcolor[HTML]{D1E1F9}0.52$\pm$.01 &
  \cellcolor[HTML]{D2E1F9}0.52$\pm$.01 &
  \cellcolor[HTML]{A3C2F3}0.74$\pm$.01 &
  \cellcolor[HTML]{A2C2F3}0.75$\pm$.01 \\ \bottomrule
\end{tabularx}
\end{adjustbox}
\caption{
Sample-level Kendall correlation between the measures analyzed in this paper, averaged across the 5~datasets in our paraphrase identification benchmark.
We report confidence intervals with bootstrap resampling.
}
\label{tab:pairwise-correlation}
\end{table*}

\subsection{Cross-lingual Results}

As discussed in Section~\ref{sec:translation-based-similarity-measures}, all three translation-based measures can be applied to both monolingual and cross-lingual sentence pairs.
The same holds for the baseline measures, even though this use case has been less prominent in previous work.
We rearrange the \pawsx{} dataset to create a cross-lingual version of the benchmark that covers 15~language pairs.
This is possible because the non-English versions of \pawsx{} are translations from the English version.
For example, we can create a cross-lingual sentence pair by pairing the English version of sentence~A with the German version of sentence~B.

Note that even though this process makes use of translation, simply reversing the translation process is not sufficient for a similarity measure to solve the resulting task, unlike with retrieval tasks that are constructed from parallel data.
As such, the cross-lingual \pawsx{} tasks are at least as hard as the original monolingual \pawsx{} tasks.

Results are reported in Table~\ref{tab:cross-lingual-pawsx-results}.
Compared to monolingual \pawsx, all measures perform worse by up to~4 points on average.
Overall, pivot translation probability is the most accurate similarity measure for cross-lingual paraphrase identification.
It should be noted, however, that both direct and pivot translation probability require the input languages to be specified, while translation cross-likelihood (like \sentencebert{} and \bertscore{}) allows them to remain unspecified.

\subsection{Correlation to Alternative Measures}

Calculating the pairwise correlation between the measures allows us to learn about similarities between the measures.
Table~\ref{tab:pairwise-correlation} visualizes the average Kendall correlations on the monolingual paraphrase identification datasets.
The translation-based measures form a cluster with a high mutual correlation, but still seem to behave differently to some degree, especially cross-likelihood.

\begin{table*}[h]
\begin{tabularx}{\textwidth}{@{}Xllll@{\hskip 20pt}llllll@{\hskip 20pt}r@{}}
\toprule
 &
  \multicolumn{4}{l}{\mbox{Individual datasets}} &
  \multicolumn{5}{@{}l}{\pawsx{} dataset} &
  \multicolumn{2}{r@{}}{\mbox{Macro-}} \\
  Language &
  \textsc{en} &
  \textsc{ru} &
  \textsc{fi} &
  \textsc{sv} &
  \textsc{de} &
  \textsc{es} &
  \textsc{fr} &
  \textsc{ja} &
  \textsc{zh} &
  \multicolumn{2}{r@{}}{\mbox{average}} \\
  Metric &
  Acc. &
  \textsc{auc} &
  \textsc{auc} &
  \textsc{auc} &
  Acc. &
  Acc. &
  Acc. &
  Acc. &
  Acc. &
  Avg. &
   \\ \midrule
\nmtscoredirect &
  72.6 &
  \underline{84.1} &
  \underline{72.4} &
  \underline{70.6} &
  \underline{73.9} &
  73.5 &
  \underline{75.7} &
  \underline{66.4} &
  \underline{68.9} &
  \underline{71.7} &
  \underline{74.3} \\
– no normalization\textsuperscript{\textdagger} &
  72.3 &
  83.3 &
  65.8 &
  67.8 &
  71.8 &
  72.8 &
  73.1 &
  62.0 &
  67.1 &
  69.3 &
  71.7 \\ \midrule
\nmtscorepivot &
  72.1 &
  \underline{84.9} &
  \underline{70.3} &
  \underline{70.9} &
  \underline{77.4} &
  \underline{76.2} &
  \underline{76.9} &
  \underline{68.4} &
  \underline{70.8} &
  \underline{74.0} &
  \underline{74.4} \\
– no normalization &
  73.0 &
  81.3 &
  64.4 &
  66.5 &
  70.2 &
  71.4 &
  70.9 &
  61.5 &
  63.5 &
  67.5 &
  70.5 \\ \midrule
\nmtscorecross &
  71.7 &
  \underline{86.6} &
  \underline{71.2} &
  \underline{72.4} &
  \underline{76.6} &
  \underline{75.1} &
  75.6 &
  \underline{65.8} &
  \underline{70.5} &
  \underline{72.7} &
  \underline{74.9} \\
– no normalization &
  71.9 &
  86.1 &
  70.6 &
  71.5 &
  75.2 &
  74.3 &
  75.5 &
  65.0 &
  69.3 &
  71.8 &
  74.4 \\ \bottomrule
\end{tabularx}
\caption{
Ablation study for the reconstruction normalizations proposed in Section~\ref{subsec:from-probability-to-similarity-score}.
The measure marked with (\textdagger) corresponds to the \textit{Prism} measure~\cite{thompson-post-2020-automatic}.
Underlined results are results that are significantly better than the other variant; in most cases reconstruction normalization leads to a significant improvement.
}
\label{tab:normalization-ablation-results}
\end{table*}

\subsection{Effect of Normalization}

Table~\ref{tab:normalization-ablation-results} presents an ablation study for the normalizations that we applied to the similarity measures (Section~\ref{subsec:from-probability-to-similarity-score}).
Overall, reconstruction normalization leads to a clear improvement.
On the English dataset, normalization does not have a positive effect on pivot translation probability and cross-likelihood.
However, since we also use English as an intermediary language for these measures, we do not think that this special case should affect our conclusions regarding the ablation study.

\section{Evaluation of Data-to-Text Generation}

We now turn to a different application of sentence similarity, namely the reference-based evaluation of data-to-text generation.
While the good performance of the translation-based measures on paraphrase identification is encouraging, this setting poses slightly different requirements on text similarity metrics.
Specifically, it is not necessary that the similarities of paraphrases and non-paraphrases are completely separable, but only that multiple hypotheses are correctly ranked with respect to a shared reference.
Moreover, it is relevant in which direction the measure is calculated.
Below we separately evaluate both directions: sim(\hypref) and sim(\refhyp), as well as their average.

\subsection{RDF-to-text}

The WebNLG 2020 challenge~\cite{ferreira2020webnlg} includes a task that requires generating natural language sentences from RDF triple sets.
Human annotators have rated system output in English and Russian using five criteria: \textit{data coverage}, \textit{relevance}, \textit{correctness}, \textit{text structure}, and \textit{fluency}.
In this paper we average the first three criteria to calculate an overall judgment of adequacy for each submitted sample (averaging first across annotators, then across individual criteria).

Since the dataset contains multiple references per RDF triple set (statistics are reported in Table~\ref{tab:dataset-statistics-webnlg}), we need to aggregate the scores computed by the automated metrics.
We follow previous work and select the maximum score across the references.

\begin{figure*}[h]
  \centering
  \includesvg[width=\linewidth,pretex=\relscale{0.8}]{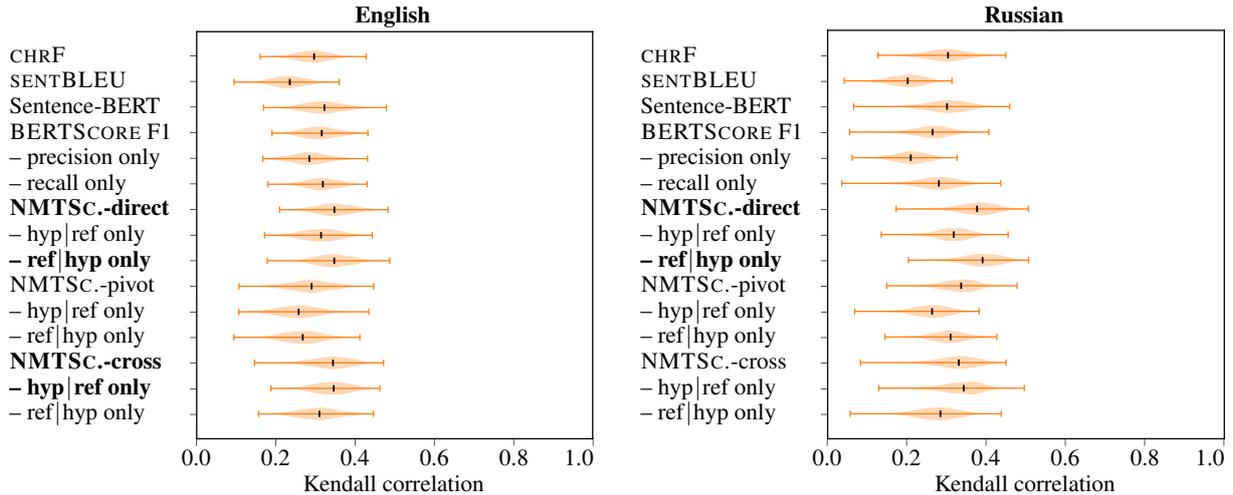}
  \caption{
  Sample-level Kendall correlation of text similarity measures to human judgments of WebNLG~2020 RDF-to-text submissions.
  A black tick denotes the correlation coefficient, and error bars denote 95\% confidence intervals.
      Measures in the top significance cluster are printed in bold.
  }
  \label{fig:webnlgu_global_kendall_results}
\end{figure*}

Figure~\ref{fig:webnlgu_global_kendall_results} shows the Kendall correlation between the similarity measures and the human judgments.
We report correlation on the level of the individual samples (also called \textit{global} correlation).
Since such meta-evaluations of metrics are known to have high statistical uncertainty, we follow~\citet{deutsch2021statistical} and estimate confidence intervals using a Boot-Both technique.
As we did before, we perform pairwise hypothesis tests and visualize the top significance cluster for each language.

Overall, direct translation probability has the highest correlation to human judgments of adequacy, indicating that translation-based measures are a competitive evaluation metric for RDF-to-text generation.

\subsection{AMR-to-text}

While the WebNLG dataset contains the output of various systems, it only encompasses two languages.
We thus complement our analysis with data collected by~\citet{fan-gardent-2020-multilingual} to evaluate a single multilingual AMR-to-text system in~15 additional European languages.
This dataset contains~50 sentences per language, with ratings by up to~10 native speakers along the criteria \textit{morphology}, \textit{word order}, \textit{semantic accuracy}, and \textit{good paraphrases}.
The languages are listed in Table~\ref{tab:dataset-statistics-amr}.

\begin{figure}[t]
  \centering
  \includesvg[width=\linewidth,pretex=\relscale{0.8}]{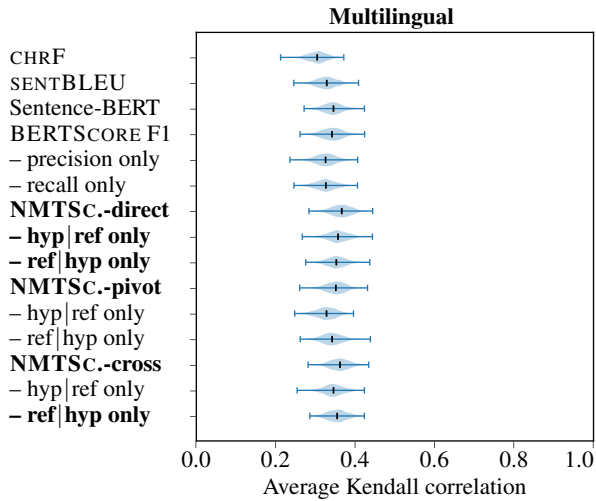}
  \caption{
    Sample-level Kendall correlation of text similarity measures to human judgments of AMR–to-text output~\cite{fan-gardent-2020-multilingual}, averaged across 15~languages.
  A black tick denotes the average correlation, and error bars denote 95\% confidence intervals.
      Measures in the top significance cluster are printed in bold.
  }
  \label{fig:amr_kendall_results}
\end{figure}

Here, we focus on semantic accuracy.
We calculate the Kendall correlation between each metric and the average human rating individually per language, and then report the average across all languages.
Figure~\ref{fig:amr_kendall_results} visualizes the confidence intervals, showing that the translation-based measures are more reliable in judging semantic accuracy than the baseline measures.

\section{Related Work}
Various strategies have been suggested to leverage translation for paraphrastic similarity.
Most related to translation cross-likelihood is perhaps the work of \citet{barzilay-mckeown-2001-extracting}, who extracted \textit{multiple translations} from a bilingual parallel corpus assuming that sentences that are aligned to the same counterpart in the other language have similar meaning.
In this paper, we revisit this idea in the context of NMT.
A different approach was pursued by \citet{bannard-callison-burch-2005-paraphrasing}, who applied phrase-based statistical MT to estimate round-trip translation probability.
This corresponds to the NMT approach of~\citet{mallinson-etal-2017-paraphrasing}, who also explore the use of multiple translation variants (\textit{multi-pivoting}) or multiple pivot languages (\textit{multilingual pivoting}).
In a variation of this approach, ~\citet{wieting-etal-2017-learning} and \citet{wieting-gimpel-2018-paranmt} used round-trip translation to generate training data for a paraphrastic sentence embedding model.

Multilingual MT allows to avoid pivoting by using zero-shot paraphrasing. This has been exploited for model analysis~\cite{tiedemann-scherrer-2019-measuring} and reference-based evaluation~\cite{thompson-post-2020-automatic}.
\citet{agrawal-etal-2021-assessing} investigate alternative techniques to estimate direct translation probability for quality estimation.
In the context of parallel corpus filtering~\cite{junczys-dowmunt-2018-dual}, \citet{chen-etal-2020-parallel-sentence} propose trie-constrained decoding to improve the efficiency of pairwise comparisons.
Future work could apply their method to the other translation-based measures.

\paragraph{Similarity of NMT representations}
An alternative line of research has used representations of NMT encoders to compare sentences, ever since it has been demonstrated that such representations can be informative~\cite{cho-etal-2014-learning, sutskever2014sequence}.
While bilingual NMT has not been found to be particularly useful for unsupervised similarity~\cite{hill-etal-2016-learning,cifka-bojar-2018-bleu}, multilingual NMT representations have proven more successful~(\citealt{schwenk-douze-2017-learning, johnson-2017-google}; among others).
However, representation learning approaches that use parallel training data without an explicit translation objective are highly competitive~\cite{wieting-etal-2019-simple,conneau-etal-2020-unsupervised,hu-etal-2021-explicit}, raising the question whether translation is indeed necessary for embedding-based similarity measures.

\section{Conclusion}
Our analysis highlights theoretical and empirical properties of translation-based text similarity measures in a multilingual setting.
Direct translation probability is the most straightforward measure (an empirical comparison of inference times is found in Appendix~\ref{sec:inference-time}).
However, it treats inputs as target sequences, and we show that accuracy on paraphrase identification can be clearly improved by normalizing with reconstruction probability.

Pivot translation probability is advantageous especially when performing cross-lingual comparisons.
Finally, translation cross-likelihood has the advantage that it achieves symmetry with a single translation direction, and that the input languages need not be specified.
The latter property also has interesting consequences for reference-based evaluation: The metric is expected to ignore whether the generated text matches the language of the reference.
This can be seen as a rigorous disentanglement of adequacy from fluency.

In comparison to baseline measures, translation-based measures are generally slower but show high accuracy on multilingual paraphrase identification, comparatively good reliability on reference-based evaluation of data-to-text generation, and little correlation to alternative measures.
Our findings thus show the usefulness of NMT translation probabilities for similarity tasks that require high attention to detail.

\section*{Limitations}
The experiments in this paper are performed on mid- and high-resource languages.
MT on low-resource languages might not yet be good enough for translation-based similarity measures to be useful.
Still, our analysis extends to more languages than previous work, including languages that have little relatedness to English.
Another limitation of translation-based text similarity measures is that the maximum sequence length supported by NMT models is often relatively short.
In Appendix~\ref{sec:dataset-statistics} we report the average character count of the text sequences used in this paper.

\section*{Acknowledgments}
This work was funded by the Swiss National Science Foundation (project MUTAMUR; no.~176727).
We would like to thank Farhad Nooralahzadeh for helpful feedback, Angela Fan for making human judgment data available, and Iker Garc{\'\i}a-Ferrero for reporting a bug in an earlier version of the library.


\bibliography{bibliography}
\bibliographystyle{acl_natbib}

\appendix

\setcounter{table}{0}
\renewcommand{\thetable}{A\arabic{table}}

\vfill
\pagebreak

\onecolumn

\section{Inference Time}\label{sec:inference-time}

\begin{table}[htb!]
\begin{tabularx}{0.4\columnwidth}{@{}Xr@{}}
\toprule
Similarity measure                 & \textit{ms} per pair \\ \midrule
\chrf                               & 0.6                             \\
\sentbleu{}                           & 0.4                             \\
\sentencebert                      & 30.7                            \\
\bertscore-F1                          & 4.5                             \\ \addlinespace[3pt]
\nmtscoredirect{}      & 22.4                            \\
– without normalization            & 12.3                            \\ \addlinespace[3pt]
\nmtscorepivot{}       & 147.8                           \\
– without normalization            & 75.2                            \\ \addlinespace[3pt]
\nmtscorecross{} & 75.0                            \\
– without normalization            & 75.0 \\ \bottomrule
\end{tabularx}
\caption{Inference time of the measures analyzed in this paper, averaged across the sentence pairs in the MRPC validation set.
We measure the average time needed to compute a measure on a sentence pair in the MRPC validation set on a RTX~2080~Ti GPU.
We use a batch size of~32 and compute the measures in both directions whenever this is required to make the measure symmetrical.
}
\label{tab:inference-time}
\end{table}

\vfill

\section{Description of Models}\label{sec:description-of-models}

\begin{table*}[htb!]
\begin{tabularx}{\textwidth}{@{}Xrrrrrrrrr@{}}
\toprule
Name  &
$N$ &
$d_{\text{model}}$ &
$d_{\text{ffn}}$  &
$h$  &
Param. &
Vocab. &
Lang. &
License &
  URL \\ \midrule
paraphrase-xlm-r-multilingual-v1 & 12 & 768  & 3072  & 12 & 278M  & 250k & 50  & Apache 2.0 & \href{https://huggingface.co/sentence-transformers/paraphrase-xlm-r-multilingual-v1}{\ExternalLink} \\
XLM-Roberta-large  (up~to~layer~17)              & 17 & 1024 & 4096  & 16 & 472M  & 250k & 100 & MIT &  \href{https://huggingface.co/xlm-roberta-large}{\ExternalLink} \\
Prism                            & 16 & 1280 & 12288 & 20 & 745M  & 64k  & 39  & MIT &  \href{https://github.com/thompsonb/prism}{\ExternalLink} \\
m2m100\_418M                     & 24 & 1024 & 4096  & 16 & 484M  & 128k & 100 & MIT &  \href{https://huggingface.co/facebook/m2m100_418M}{\ExternalLink} \\
m2m100\_1.2B                     & 48 & 1024 & 8192  & 16 & 1239M & 128k & 100 & MIT &  \href{https://huggingface.co/facebook/m2m100_1.2B}{\ExternalLink} \\ \bottomrule
\end{tabularx}
\caption{Hyperparameters of the Transformer models used in this paper, as well as number of parameters, vocabulary size and number of supported languages.}
\label{tab:model-sizes}
\end{table*}

\vfill

\section{Metric Version Signatures}

\chrf: {\footnotesize\texttt{nrefs:1|case:mixed|eff:yes|nc:6|nw:0|space:no|version:2.0.0}}

\medskip \noindent
\sentbleu: {\footnotesize\texttt{nrefs:1|case:mixed|eff:yes|tok:13a|smooth:exp|version:2.0.0}} \\
– \textsc{ja}: {\footnotesize\texttt{nrefs:1|case:mixed|eff:yes|tok:ja-mecab-0.996-IPA|smooth:exp|version:2.0.0}} \\
– \textsc{zh}: {\footnotesize\texttt{nrefs:1|case:mixed|eff:yes|tok:zh|smooth:exp|version:2.0.0}}

\medskip \noindent
\bertscore: {\footnotesize\texttt{xlm-roberta-large\_L17\_no-idf\_version=0.3.11(hug\_trans=4.17.0)}}

\medskip \noindent
\nmtscore: \\
{\footnotesize\texttt{NMTScore-direct|model:prism|normalized|both-directions|v0.2.0|hf4.17.0}} \\
{\footnotesize\texttt{NMTScore-pivot|pivot-lang:en|model:prism|normalized|both-directions|v0.2.0|hf4.17.0}} \\
{\footnotesize\texttt{NMTScore-cross|tgt-lang:en|model:prism|normalized|both-directions|v0.2.0|hf4.17.0}}

\vfill
\pagebreak

\section{Dataset Statistics}\label{sec:dataset-statistics}

\vfill

\begin{table*}[htb!]
\begin{tabularx}{\textwidth}{@{}XXrrrllr@{}}
\toprule
 & Split & \multicolumn{1}{l}{Positive pairs} & \multicolumn{1}{l}{Negative pairs} & \multicolumn{1}{l}{Avg. chars} & License & Domains & URL \\ \midrule
\textsc{en} & Validation & 239   & 128  & 109 & unspecified   & news            & \href{https://huggingface.co/datasets/glue}{\ExternalLink}   \\
   & Test   & 1002  & 566  & 107 &               &                 &                                  \\ \midrule
\textsc{ru} & Test   & 1152  & 772  & 60  & MIT License   & news            & \href{http://paraphraser.ru}{\ExternalLink}                \\ \midrule
\textsc{fi} & Test   & 15368 & 5574 & 73  & CC-BY-SA 4.0  & subtitles, news & \href{https://turkunlp.org/paraphrase.html}{\ExternalLink}   \\ \midrule
\textsc{sv} & Test   & 783   & 298  & 45  & CC-BY-SA 4.0  & subtitles       & \href{https://turkunlp.org/paraphrase.html}{\ExternalLink}   \\ \midrule
\textsc{de} & Validation & 831   & 1101 & 119 & public domain & wikipedia       & \href{https://huggingface.co/datasets/paws-x}{\ExternalLink} \\
   & Test   & 895   & 1073 & 121 &               &                 &                                        \\
\textsc{es} & Validation & 847   & 1115 & 117 &               &                 &                                        \\
   & Test   & 907   & 1092 & 118 &               &                 &                                        \\
\textsc{fr} & Validation & 860   & 1132 & 120 &               &                 &                                        \\
   & Test   & 903   & 1083 & 121 &               &                 &                                        \\
\textsc{ja} & Validation & 854   & 1126 & 58  &               &                 &                                        \\
   & Test   & 883   & 1063 & 60  &               &                 &                                        \\
\textsc{zh} & Validation & 853   & 1131 & 43  &               &                 &                                        \\
   & Test   & 894   & 1081 & 44  &               &                 &                                        \\ \bottomrule
\end{tabularx}
\caption{Dataset statistics for our multilingual paraphrase identification benchmark.}
\label{tab:dataset-statistics}
\bigskip
\end{table*}

\vfill

\begin{table*}[htb!]
\begin{tabularx}{\textwidth}{@{}Xrr@{\hskip 40pt}rr@{}}
\toprule
 & \multicolumn{1}{l}{Validation}     & \multicolumn{1}{l}{}               & \multicolumn{1}{l}{Test}           & \multicolumn{1}{l}{}               \\
 & \multicolumn{1}{l}{Positive pairs} & \multicolumn{1}{l@{\hskip 40pt}}{Negative pairs} & \multicolumn{1}{l}{Positive pairs} & \multicolumn{1}{l}{Negative pairs} \\  \midrule
\textsc{en+de} & 1662 & 2202 & 1790 & 2146 \\
\textsc{en+es} & 1694 & 2230 & 1814 & 2184 \\
\textsc{en+fr} & 1720 & 2264 & 1806 & 2166 \\
\textsc{en+ja} & 1708 & 2252 & 1766 & 2126 \\
\textsc{en+zh} & 1706 & 2262 & 1788 & 2162 \\
\textsc{de+es} & 1640 & 2168 & 1790 & 2146 \\
\textsc{de+fr} & 1658 & 2194 & 1788 & 2132 \\
\textsc{de+ja} & 1646 & 2184 & 1748 & 2094 \\
\textsc{de+zh} & 1648 & 2190 & 1772 & 2126 \\
\textsc{es+fr} & 1688 & 2220 & 1806 & 2166 \\
\textsc{es+ja} & 1678 & 2208 & 1766 & 2124 \\
\textsc{es+zh} & 1674 & 2218 & 1788 & 2160 \\
\textsc{fr+ja} & 1702 & 2242 & 1764 & 2114 \\
\textsc{fr+zh} & 1702 & 2252 & 1784 & 2142 \\
\textsc{ja+zh} & 1688 & 2240 & 1744 & 2104 \\ \bottomrule
\end{tabularx}
\caption{Dataset statistics for the cross-lingual \pawsx{} benchmark.}
\label{tab:cross-lingual-pawsx-dataset-statistics}
\end{table*}

\vfill
\pagebreak

\begin{table*}[htb!]
\begin{tabularx}{\textwidth}{@{}Xrrrrr@{}}
\toprule
Language &
  Documents &
  Systems &
  Samples &
  Avg. references &
  Avg. reference characters \\ \midrule
\textsc{en} &
  178 &
  16 &
  2848 &
  2.9 &
  132 \\
\textsc{ru} &
  110 &
  7 &
  770 &
  2.5 &
  123 \\ \bottomrule
\end{tabularx}
\caption{Statistics for the WebNLG 2020 RDF-to-text dataset of human judgments.}
\label{tab:dataset-statistics-webnlg}
\bigskip
\end{table*}

\begin{table*}[htb!]
\begin{tabularx}{\textwidth}{@{}Xrrrrrrrrrrrrrrr@{}}
\toprule
Language &
  \textsc{da} &
  \textsc{el} &
  \textsc{es} &
  \textsc{fi} &
  \textsc{it} &
  \textsc{nl} &
  \textsc{pt} &
  \textsc{sv} &
  \textsc{bg} &
  \textsc{cs} &
  \textsc{et} &
  \textsc{hu} &
  \textsc{lv} &
  \textsc{pl} &
  \textsc{ro} \\ \midrule
Samples &
  50 &
  50 &
  50 &
  50 &
  50 &
  50 &
  50 &
  50 &
  50 &
  50 &
  50 &
  50 &
  50 &
  50 &
  50 \\
Avg. ref. chars &
  56 &
  61 &
  60 &
  56 &
  62 &
  61 &
  59 &
  52 &
  130 &
  129 &
  42 &
  138 &
  129 &
  151 &
  130 \\ \bottomrule
\end{tabularx}
\caption{Statistics for the multilingual AMR–to-text dataset of human judgments.}
\label{tab:dataset-statistics-amr}
\end{table*}

\vfill

\section{Other NMT Models}\label{sec:other-nmt-models}

\begin{table*}[htb!]
\begin{tabularx}{\textwidth}{@{}Xllll@{\hskip 20pt}llllll@{\hskip 20pt}r@{}}
\toprule
 &
  \multicolumn{4}{l}{\mbox{Individual datasets}} &
  \multicolumn{5}{@{}l}{\pawsx{} dataset} &
  \multicolumn{2}{r@{}}{\mbox{Macro-}} \\
  Language &
  \textsc{en} &
  \textsc{ru} &
  \textsc{fi} &
  \textsc{sv} &
  \textsc{de} &
  \textsc{es} &
  \textsc{fr} &
  \textsc{ja} &
  \textsc{zh} &
  \multicolumn{2}{r@{}}{\mbox{average}} \\
  Metric &
  Acc. &
  \textsc{auc} &
  \textsc{auc} &
  \textsc{auc} &
  Acc. &
  Acc. &
  Acc. &
  Acc. &
  Acc. &
  Avg. &
   \\ \midrule
\multicolumn{12}{@{}l}{m2m100\_418M} \\ \addlinespace[3pt]
\nmtscoredirect & 72.0 & 83.2 & 71.1 & 71.1 & 71.1 & 69.0 & 72.3 & 61.9 & 65.4 & 67.9 & 73.1 \\
\nmtscorepivot & 72.4 & 84.2 & 68.2 & 70.3 & 73.2 & 72.0 & 72.2 & 64.0 & 67.9 & 69.9 & 73.0 \\
\nmtscorecross & 72.1 & 85.1 & 69.7 & 71.5 & 71.6 & 72.6 & 72.2 & 63.1 & 66.7 & 69.2 & 73.5 \\
\midrule
\multicolumn{12}{@{}l}{m2m100\_1.2B} \\ \addlinespace[3pt]
\nmtscoredirect & 72.9 & 84.0 & 71.4 & 71.2 & 73.0 & 70.2 & 72.4 & 62.4 & 66.4 & 68.9 & 73.7 \\
\nmtscorepivot & 74.1 & 84.5 & 69.1 & 69.6 & 75.1 & 73.0 & 73.3 & 65.8 & 70.2 & 71.5 & 73.8 \\
\nmtscorecross & 72.8 & 85.0 & 70.0 & 71.0 & 74.1 & 73.0 & 73.3 & 66.2 & 69.5 & 71.2 & 74.0 \\
    \bottomrule
\end{tabularx}
\caption{
Comparison of translation-based text similarity measures when using two other multilingual NMT models~(\textit{M2M-100};~\citealp{fan2021beyond}).
Overall, the accuracy of all three measures is slightly lower compared to the \prism{} NMT model but still competitive compared to the embedding baselines.
}
\label{tab:m2m100-results}
\end{table*}

\begin{table*}[htb!]
\setlength{\tabcolsep}{3.5pt}
\begin{adjustbox}{width=\textwidth,center}
\begin{tabularx}{1.135\textwidth}{@{}Xcccccccccccccccr@{}}
\toprule
 &
  \textsc{en} &
  \textsc{en} &
  \textsc{en} &
  \textsc{en} &
  \textsc{en} &
  \textsc{de} &
  \textsc{de} &
  \textsc{de} &
  \textsc{de} &
  \textsc{es} &
  \textsc{es} &
  \textsc{es} &
  \textsc{fr} &
  \textsc{fr} &
  \textsc{ja} &
   \\ \addlinespace[-6pt]
 &
  + &
  + &
  + &
  + &
  + &
  + &
  + &
  + &
  + &
  + &
  + &
  + &
  + &
  + &
  + &
  Avg.\\ \addlinespace[-5pt]
 &
  \textsc{de} &
  \textsc{es} &
  \textsc{fr} &
  \textsc{ja} &
  \textsc{zh} &
  \textsc{es} &
  \textsc{fr} &
  \textsc{ja} &
  \textsc{zh} &
  \textsc{fr} &
  \textsc{ja} &
  \textsc{zh} &
  \textsc{ja} &
  \textsc{zh} &
  \textsc{zh} &
   \\ \midrule
\multicolumn{12}{@{}l}{m2m100\_418M} \\ \addlinespace[3pt]
\nmtscoredirect{} & 72.5 & 70.9 & 72.2 & 63.3 & 65.0 & 67.9 & 69.0 & 61.0 & 63.1 & 68.4 & 60.6 & 62.0 & 61.4 & 63.2 & 60.6 & 65.4 \\
\nmtscorepivot{} & 73.8 & 73.1 & 73.6 & 63.9 & 65.4 & 71.9 & 70.5 & 63.0 & 63.8 & 70.1 & 62.8 & 64.3 & 62.7 & 63.5 & 62.0 & 67.0 \\
\nmtscorecross & 72.8 & 72.3 & 72.2 & 61.9 & 62.3 & 69.7 & 69.0 & 60.2 & 63.2 & 69.8 & 61.5 & 61.5 & 60.5 & 61.7 & 61.9 & 65.4 \\ \midrule
\multicolumn{12}{@{}l}{m2m100\_1.2B} \\ \addlinespace[3pt]
\nmtscoredirect{} & 75.0 & 72.4 & 73.0 & 64.8 & 67.0 & 71.4 & 71.7 & 62.7 & 65.1 & 69.8 & 61.5 & 63.4 & 62.7 & 65.1 & 62.6 & 67.2 \\
\nmtscorepivot{} & 75.9 & 74.0 & 74.4 & 66.4 & 67.5 & 72.3 & 72.4 & 64.4 & 66.5 & 70.9 & 63.4 & 65.6 & 64.1 & 64.9 & 63.4 & 68.4 \\
\nmtscorecross & 74.8 & 74.1 & 73.8 & 63.0 & 63.6 & 70.9 & 70.5 & 61.4 & 63.5 & 71.3 & 61.6 & 62.5 & 61.8 & 63.8 & 63.3 & 66.7 \\ \bottomrule
\end{tabularx}
\end{adjustbox}
\caption{
Comparison of translation-based text similarity measures on the cross-lingual \pawsx{} dataset, using two other multilingual NMT models~(\textit{M2M-100};~\citealp{fan2021beyond}).
Again, the average accuracy is lower compared to the \prism{} NMT model that we used for the main experiments, but superior to the baselines.
}
\label{tab:m2m100-cross-lingual-pawsx-results}
\end{table*}

\vfill
\pagebreak

\section{MRPC Examples}\label{sec:examples}

\begin{table*}[htb!]
\begin{tabularx}{\textwidth}{@{}Xrrr@{}}
\toprule
\textbf{Sentence Pair} &
  \textbf{Gold} &
  \textbf{\phantom{-------}\textsc{SBERT}} &
  \textbf{\nmtscore{}} \\ \midrule
\textit{The Dow Jones Industrial Average fell 0.7 per cent to 9,547.43 while the S\&P 500 was 0.8 per cent weaker at 1,025.79.} &
  0 &
  0.93 &
  0.21 \\ \addlinespace[3pt]
  \textit{The Dow Jones industrial average fell 44 points, or 0.46 percent, to 9,568.} \\ \midrule
\textit{So far, they have searched Pennsylvania, Ohio, Michigan, Illinois and Indiana, authorities in those state said.} &
  0 &
  0.81 &
  0.08 \\ \addlinespace[3pt]
  \textit{So far, authorities also have searched areas in Pennsylvania, Ohio, Indiana, and Michigan.} \\ \midrule
\textit{MEN who drink tea, particularly green tea, can greatly reduce their risk of prostate cancer, a landmark WA study has found.} &
  1 &
  0.91 &
  0.14 \\ \addlinespace[3pt]
  \textit{DRINKING green tea can dramatically reduce the risk of men contracting prostate cancer, a study by Australian researchers has discovered.} \\ \midrule
\textit{Bashir felt he was being tried by opinion not on the facts, Mahendradatta told Reuters.} &
  1 &
  0.87 &
  0.18 \\ \addlinespace[3pt]
  \textit{Bashir also felt he was being tried by opinion rather than facts of law, he added.} \\ \bottomrule
\end{tabularx}
\caption{MRPC examples with a high disagreement between \sentencebert{} cosine similarity and \nmtscorecross{}.}
\label{tab:sentencebert-examples}
\end{table*}

\begin{table*}[htb!]
\begin{tabularx}{\textwidth}{@{}Xrrr@{}}
\toprule
\textbf{Sentence Pair} &
  \textbf{Gold} &
  \textbf{\bertscore{}} &
  \textbf{\nmtscore{}} \\ \midrule
\textit{Batters faced: Sheets 28, Vizcaino 2, DeJean 4, Clement 26, Alfonseca 4, Guthrie 2, Farnsworth 4.} &
  0 &
  0.50 &
  0.07 \\ \addlinespace[3pt]
  \textit{Batters faced: Franklin 25, Kieschnick 7, Foster 2, Leskanic 3, DeJean 4, Prior 28, Alfonseca 2, Guthrie 2, Cruz 7, Remlinger 6.} \\ \midrule
\textit{But the technology-laced Nasdaq Composite Index was up 5.91 points, or 0.35 percent, at 1,674.35.} &
  0 &
  0.45 &
  0.07 \\ \addlinespace[3pt]
  \textit{The broader Standard \& Poor's 500 Index .SPX was off 1.07 points, or 0.11 percent, at 1,010.59.} \\ \midrule
\textit{They also found shortness was associated with a family history of hearing loss.} &
  1 &
  0.46 &
  0.09 \\ \addlinespace[3pt]
  \textit{Shortness was found twice as often in those with hearing loss.} \\ \midrule
\textit{Kollar-Kotelly has scheduled another antitrust settlement compliance hearing for January.} &
  1 &
  0.41 &
  0.08 \\ \addlinespace[3pt]
  \textit{The judge scheduled another oversight hearing for late January.} \\ \bottomrule
\end{tabularx}
\caption{MRPC examples with a high disagreement between rescaled \bertscore{}-F1 and \nmtscorecross{}.}
\label{tab:bertscore-examples}
\end{table*}

\vfill
\pagebreak

\section{Cross-lingual \pawsx{} Examples}\label{sec:cross-lingual-pawsx-examples}

\begin{table*}[htb!]
\begin{tabularx}{\textwidth}{@{}Xrrr@{}}
\toprule
\textbf{Sentence Pair} &
  \textbf{Gold} &
  \textbf{\phantom{-------}\textsc{SBERT}} &
  \textbf{\nmtscore{}} \\ \midrule
\textsc{en}: \textit{Write once~, run anywhere} &
  0 &
  0.76 &
  0.21 \\ \addlinespace[3pt]
  \textsc{fr}: \textit{Écrivez n'importe où, une fois exécuté} \\ \midrule
\textsc{en}: \textit{Worcester is a town and county city of Worcestershire in England~.} &
  0 &
  0.93 &
  0.37 \\ \addlinespace[3pt]
  \textsc{de}: \textit{Worcestershire ist eine Stadt und Kreisstadt von Worcester, England.} \\ \midrule
\textsc{en}: \textit{The Jiul de Vest River is a tributary of the Jidanul River in Romania~.} &
  0 &
  0.78 &
  0.30 \\ \addlinespace[3pt]
  \textsc{de}: \textit{Der Jidanul ist ein Nebenfluss des Jiul de Vest, Rumänien.} \\ \midrule
\textsc{en}: \textit{The Cugir River is a tributary of the Ghișag River in Romania~.} &
  1 &
  0.89 &
  0.48 \\ \addlinespace[3pt]
  \textsc{de}: \textit{Der Fluss Cugir ist ein Nebenfluss des Ghiaag in Rumänien.} \\ \midrule
\textsc{en}: \textit{Film stars Lily Rabe~, Timothée Chalamet~, Lili Reinhart~, Anthony Quintal~, Oscar Nunez and Rob Huebel .} &
  1 &
  0.80 &
  0.49 \\ \addlinespace[3pt]
  \textsc{fr}: \textit{Le film met en vedette Oscar Nunez, Rob Huebel, Timothée Chalamet, Lily Rabe, Anthony Quintal et Lili Reinhart.} \\ \bottomrule
\end{tabularx}
\caption{Cross-lingual \pawsx{} examples with a high disagreement between \sentencebert{} cosine similarity and \nmtscorepivot{}.}
\label{tab:sentencebert-examples-crosslingual-pawsx}
\end{table*}

\end{document}